\documentclass[runningheads]{llncs}
\usepackage[utf8]{inputenc}

\usepackage{graphicx}
\usepackage[export]{adjustbox}  
\usepackage{multirow}
\usepackage{booktabs}
\usepackage{amsmath}
\usepackage{amssymb}
\usepackage[sort, numbers]{natbib}
\usepackage{svg}
\usepackage{subcaption}
\usepackage{hyperref}
\usepackage{tabularx}
\usepackage{xfp}
\usepackage{siunitx}
\usepackage{array}
\newcolumntype{L}[1]{>{\raggedright\let\newline\\\arraybackslash\hspace{0pt}}m{#1}}
\newcolumntype{C}[1]{>{\centering\let\newline\\\arraybackslash\hspace{0pt}}m{#1}}
\newcolumntype{R}[1]{>{\raggedleft\let\newline\\\arraybackslash\hspace{0pt}}m{#1}}

\usepackage{orcidlink}

\usepackage[misc,geometry]{ifsym}
\begin{document}
\setcitestyle{square}
\title{Long-Range Transformer Architectures for Document Understanding}

\titlerunning{Long-Range Transformer Architectures for Document Understanding}

\author{\mbox{Thibault Douzon\inst{1,2} \orcidlink{0009-0001-1649-5373}} \and \mbox{Stefan Duffner\inst{1} \orcidlink{0000-0003-0374-3814}} \and \mbox{Christophe Garcia\inst{1} \orcidlink{0000-0001-7997-9837}} \and \mbox{Jérémy Espinas\inst{2}}}

\authorrunning{T. Douzon et al.}

\institute{INSA Lyon, LIRIS, Lyon, France \\ \email{firstname.lastname@insa-lyon.fr}
\and Esker, Lyon, France \\ \email{firstname.lastname@esker.com}
}
%
%
%
%
\maketitle              
\begin{abstract}
Since their release, Transformers have revolutionized many fields from Natural Language Understanding to Computer Vision.
Document Understanding (DU) was not left behind with first Transformer based models for DU dating from late 2019.
However, the computational complexity of the self-attention operation limits their capabilities to small sequences.
In this paper we explore multiple strategies to apply Transformer based models to long multi-page documents.
We introduce 2 new multi-modal (text + layout) long-range models for DU.
They are based on efficient implementations of Transformers for long sequences.
Long-range models can process whole documents at once effectively and are less impaired by the document's length.
We compare them to LayoutLM, a classical Transformer adapted for DU and pre-trained on millions of documents.
We further propose 2D relative attention bias to guide self-attention towards relevant tokens without harming model efficiency.
We observe improvements on multi-page business documents on Information Retrieval for a small performance cost on smaller sequences.
Relative 2D attention revealed to be effective on dense text for both normal and long-range models.

\end{abstract}
\section{Introduction}

Digital documents are everywhere around us, in the form of born digital PDF or scanned paper, they carry much information and can be easily exchanged.
They can be used to exchange information from an issuer to its recipient or to archive its content. 
Information is generally structured in some way depending on the document type.
Invoices and scientific articles, for example, do not follow the same structure because their objective is different.
Both are codified to carry information very efficiently  such that most invoices and most articles look the same but differ in content.
Document Understanding is a field gaining increasing attention in the past years, as automating document-related processes can drastically improve efficiency of information processing in a wide range of fields and industries.
Recent advances in Neural Network architectures allowed better document understanding and enabled tackling more complex tasks: Question Answering~\cite{Mathew2021}, Layout Segmentation~\cite{Li2020} and Information Extraction~\cite{Huang2019}.

In particular, models based on Transformer architectures have led to a breakthrough in these domains since their first release in 2017~\cite{Vaswani2017}. 
They have been widely used on Natural Language Processing tasks and their performance is unequaled~\cite{Wang2018}. 
The implementation of Transformer models in DU~\cite{Xu2020} was swift, revealing the potential of attention-based models in this domain.
More recently, the trend is towards multi-modal models that combine multiple different information representations such as text, image, sound, layout.
Those models have shown great results on short, single page documents, but are difficult to apply to long, multi-page or dense documents.
This is because the self-attention 
time and space computational complexity is $O(N^2)$ where $N$ is the length of the sequence.
It effectively limits the usage of Transformer models on long sequences due to either long training or lack of GPU memory.

In this work, we explore several approaches and architectures in order to use Transformer models on long documents. 
For simplicity, we limit our study to text and layout (i.e.,\ text position in the page) modalities, and chose to focus on document length to evaluate the model efficiency.
We compare various encoder-only models on Sequence Tagging tasks with business and academic documents. 
We also study the impact of relative attention based on document layout instead of a linear token position, and its implementation for long-range Transformers.

\section{Related Work}

\subsection{From NLP to Document Understanding}
This work derives from both long-range Transformers proposed in NLP tasks, trying to process longer sequences at once and Transformer architectures adapted to DU.
Before the proposal of Transformers, the {\it de facto} architecture for NLP has been Recurrent Neural Networks. 
Multiple improvements have been proposed, for example to tackle vanishing gradients like Long-Short Term Memory cells~\cite{Hochreiter1997}. 
Coupled with Conditional Random Fields, bidirectional LSTM encoders were then capable at most text understanding task~\cite{Lample2016}.
For more complex Information Retrieval, where target information can span multiple tokens, \texttt{BIESO} tags allow better decoding by precisely locating the beginning and end of the information.
Although long sequences can be processed with Recurrent Neural Networks, longer input negatively affects the performance of encoder-decoder architectures~\cite{Bahdanau2016}. 
Hence, the attention mechanism was quickly adopted for those architectures as an ``information highway'' between the encoder and the decoder.

In addition to these new architecture developments, large progress has been made in the past years on how to learn usable word representations.
Before, word embeddings were trained at the same time as the other model's parameters. 
Then, approaches like Word2Vec and GloVe~\cite{Mikolov2013, Pennington2014} showed that self-supervised learning improves finetuning on all tasks.
Major improvements came from contextual embeddings, first introduced by Elmo~\cite{Peters2018}.
Contrary to static embeddings, contextual embeddings can better represent words with multiple meanings in adequation with their surroundings.

This is where Transformer models rose, heavily relying on (self-)attention and pre-training giving unprecedented performance at the time.
Most NLP challenges leaderboards were monopolized by BERT like models, growing bigger and deeper by the day~\cite{Liu2019, Brown2020, chowdheryPaLMScalingLanguage2022}.

In parallel to those quick improvements, the DU community developped alternatives to bi-LSTM, using multiple modalities to provide more useful information to the model. 
Some used convolutions over a grid mixing image and text~\cite{Katti2018, Denk2019}, others proposed graph-based models~\cite{Yu2020} to represent a document.


The revolution in DU came from Transformer architectures. 
Pre-trained models able to leverage large document collections outperformed all previous approaches. 
LayoutLM~\cite{Xu2020}, for example, only introduced 2D positional embeddings over BERT and was pre-trained on the RVL-CDIP~\cite{Lewis2006} collection. 
It opened the way to many other models applying Transformers to previous design~\cite{Denk2019}, leveraging end-to-end capacities of encoder-decoder models~\cite{Powalski2021}, or providing image and text to the models like a visual Transformer~\cite{Kolesnikov2021}.
Because the Transformer output is independent of the sequence order, positional embeddings are classically added to the input.
It is also possible to introduce relative bias to the self-attention mechanism to promote local interactions inside the self-attention.

Most recent models for DU propose to leverage as much information as possible by using multiple modalities: text, layout and image.
Either by combining Convolutional Neural Networks with Transformers~\cite{Xu2021, Powalski2021} or mixing visual with classical Transformers~\cite{Huang2022, Kim2022}.
Even though those approches provide superior results, we chose to not include image information to our architectures.

\subsection{Long Range Transformers}

Since the introduction of BERT~\cite{Devlin2019} and GPT~\cite{Radford2019}, Transformers have demonstrated their capacity to understand and model language~\cite{Wang2018}.
Their ability to manipulate words can be visualised through the amount of attention each token allows to other tokens.
However, dot-product attention computation involves a $O(N^2)$ time and memory complexity where $N$ is the sequence length.
It limits the capacity of Transformer-based models in dealing with long sequences as they need too much GPU memory and/or take too long to process.

Many modifications have been proposed to replace the attention layer with some efficient approximation that can be computed in $O(N)$ or $O(N\log(N))$.
They have been developped and tested with NLP tasks where long sequences are most likely to be found like long text summarization and translation.
Some models use attention patterns~\cite{Beltagy2020, Zaheer2020, Ainslie2020} to limit token attention to a fixed number of other tokens. Some combination of sliding window, global and random patterns provide a simple but efficient attention.
A balance needs to be found between more attention context and attention complexity.
It is also possible to learn the attention pattern by limiting attention to tokens that share some locality sensitive hash~\cite{Kitaev2020}.
Others proposed to replace the $N \times N$ attention matrix with a low-rank approximation. 
Empirical observations on multiple NLP tasks show that the attention matrix can be replaced with a lower rank approximation without harming the attention process too much~\cite{Wang2020}.

However, long range Transformer architectures have not yet been used on DU tasks, mostly due to datasets not containing lengthy documents.

\section{Datasets}

We used 2 document datasets, where our choice was mainly made based on document length and the task itself. We wanted a NLP task that can be represented as Sequence Tagging in order to test the whole encoder with long inputs.
Both datasets consist of English-only documents with close to perfect OCR extraction. 
They provide word-level axis aligned bounding boxes in the form that can be fed to the model as layout information.
We use the OCR provided order for the input sequence and do not further analyze documents to extract their structure.

\subsection{Business Documents}
The first dataset consists of Customer Orders submitted to a commercial platform between 2018 and 2021. Due to privacy concerns, these documents cannot be shared. 
It contains 80k documents that can be divided in 9000 different issuers with no more than 50 documents from the same issuer.
Usually, an issuer only emits documents with the same template for convenience.
About 55\% of documents can be tokenized into a sequence of 512 tokens which fit into classical Transformer default maximum length.
Only 5\% of documents are longer than 2048 tokens, following a long tail of distribution.
In order to evaluate the models' generalization abilities, we split into train, validation and test sets such that templates in the test set have not been seen by the model during training.

\begin{figure}[t]
    \centering
    \begin{subfigure}[t]{0.40\textwidth}
        \includegraphics[max height=7cm]{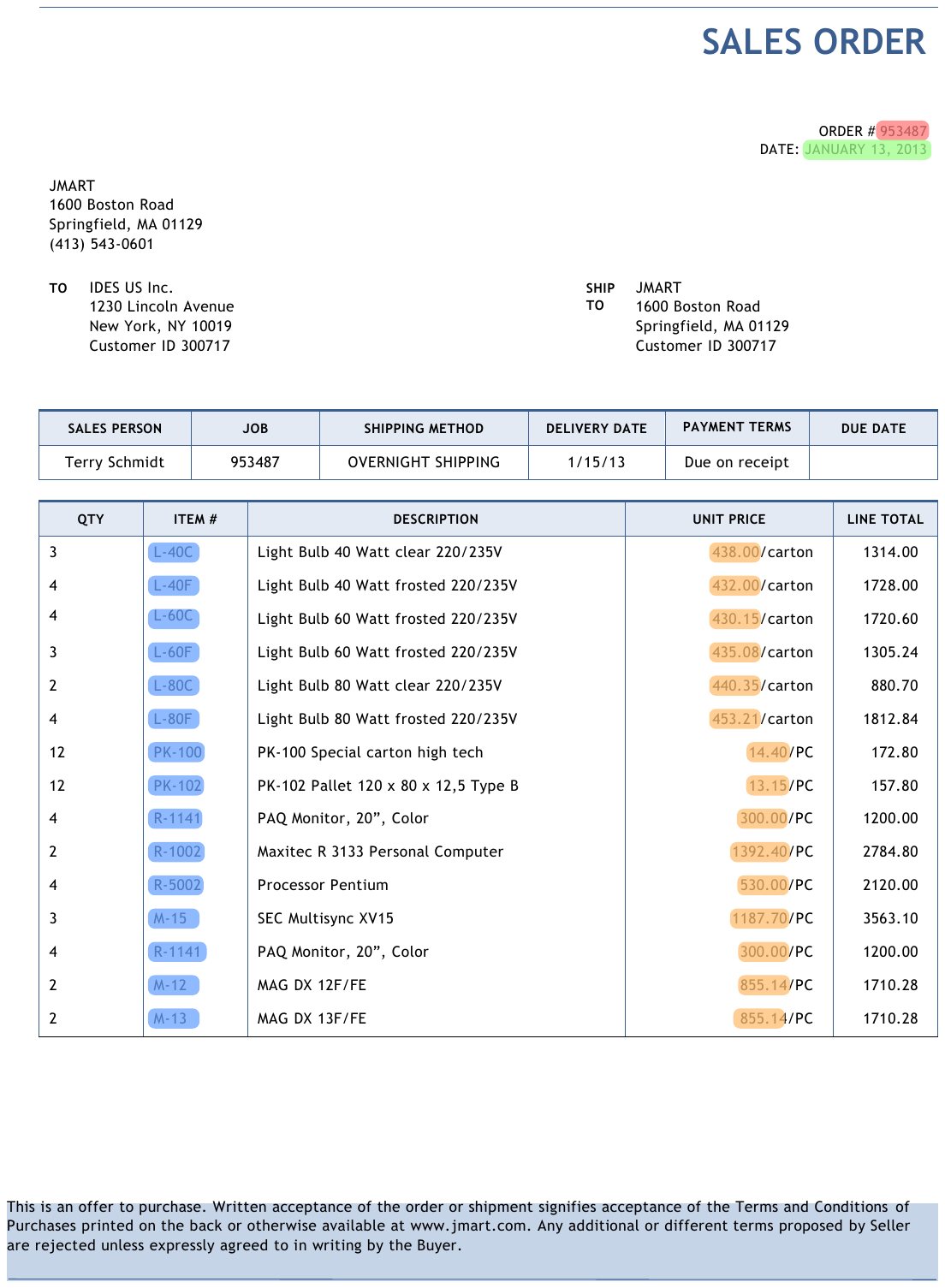}
    \end{subfigure}
    \hfill
    \begin{subfigure}[t]{0.40\textwidth}
        \includegraphics[max height=7cm]{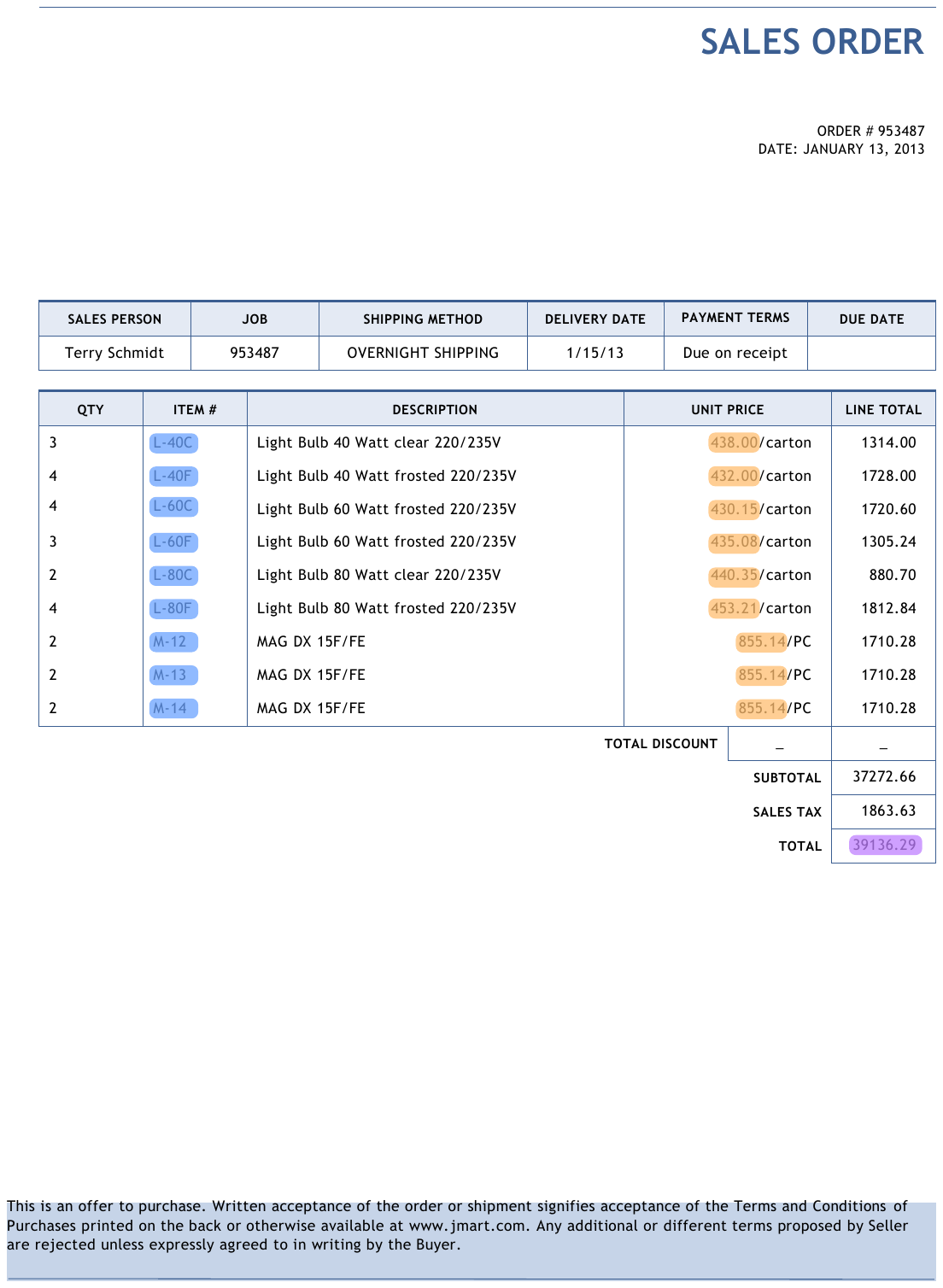}
    \end{subfigure}
    \caption{Sample pages with colored labels similar to those in the Business Documents dataset. Both pages come from the same document, the first page in on the left and the last page on the right. Some information are repeated across pages of a document.}
    \label{fig:order_sample}
\end{figure}

The task consists of Information Extraction on multiple known classes: \textit{document number}, \textit{date}, \textit{total amount}, \textit{item ID numbers} and \textit{item quantities}.
Some information only appears once in the document (e.g.,\ \textit{document number}, \textit{date} and \textit{total amount}) while others are repeated for each line item in the business order.
We call \textit{header} fields those only occurring once and \textit{table} fields others as they are most of the time structured in a table layout.
There could be between 1 and 50 items present in any document, their number is not known in advance.
\figurename~\ref{fig:order_sample} shows the labeling of a multi-page document. 
Even though header field are sometimes repeated on each page, it is only labeled once in order to stay consistent acrosstemplates.
Labels are provided at the word level based on manual customer document extraction. 
We also controlled labeling quality and rejected from the dataset documents with missing mandatory fields or wrong number of line items.

A superset of this dataset was used for pre-training models on business documents. It consists of 300k Customer Orders and 100k Invoices from the same commercial platform.
All documents were submitted and processed by the platform but later rejected due to labeling errors or bad habits.
Fortunately, this does not impact the OCR quality and allows us to pre-train our models on a large collection of recent documents.
We chose to use it for pre-training instead of RVL-CDIP~\cite{Harley2015} for the OCR quality difference.

\subsection{DocBank}
DocBank~\cite{Li2020} is a dataset containing 500k public research article pages.
It contains English documents spanning various research fields.
Documents were obtained on arXiv and were annotated with PDFPlumber, a PDF parser that accurately extracts item bounding boxes.
The task consists in document layout analysis. \citet{Li2020} provide both pixel and word-level annotations for CV and NLP models.
The order of words is defined from top-to-bottom and left-to-right, except for multicolumn documents where whole columns are ordered left-to-right.
In this work we will only use textual information along the word 2D positions.

Docbank segmentation task contains 12 categories (e.g.\ \textit{title}, \textit{paragraph}, \textit{figure} etc.) representing semantic parts of a research article.
Because articles contain dense paragraphs, most pages are longer than 512 tokens once tokenized. 
In fact only 11\% of the test documents contains less than 512 tokens and 84\% contains between 512 and 2048 tokens.

\begin{figure}[t]
    \centering
    \begin{subfigure}[t]{0.40\textwidth}
        \includegraphics[max height=7cm]{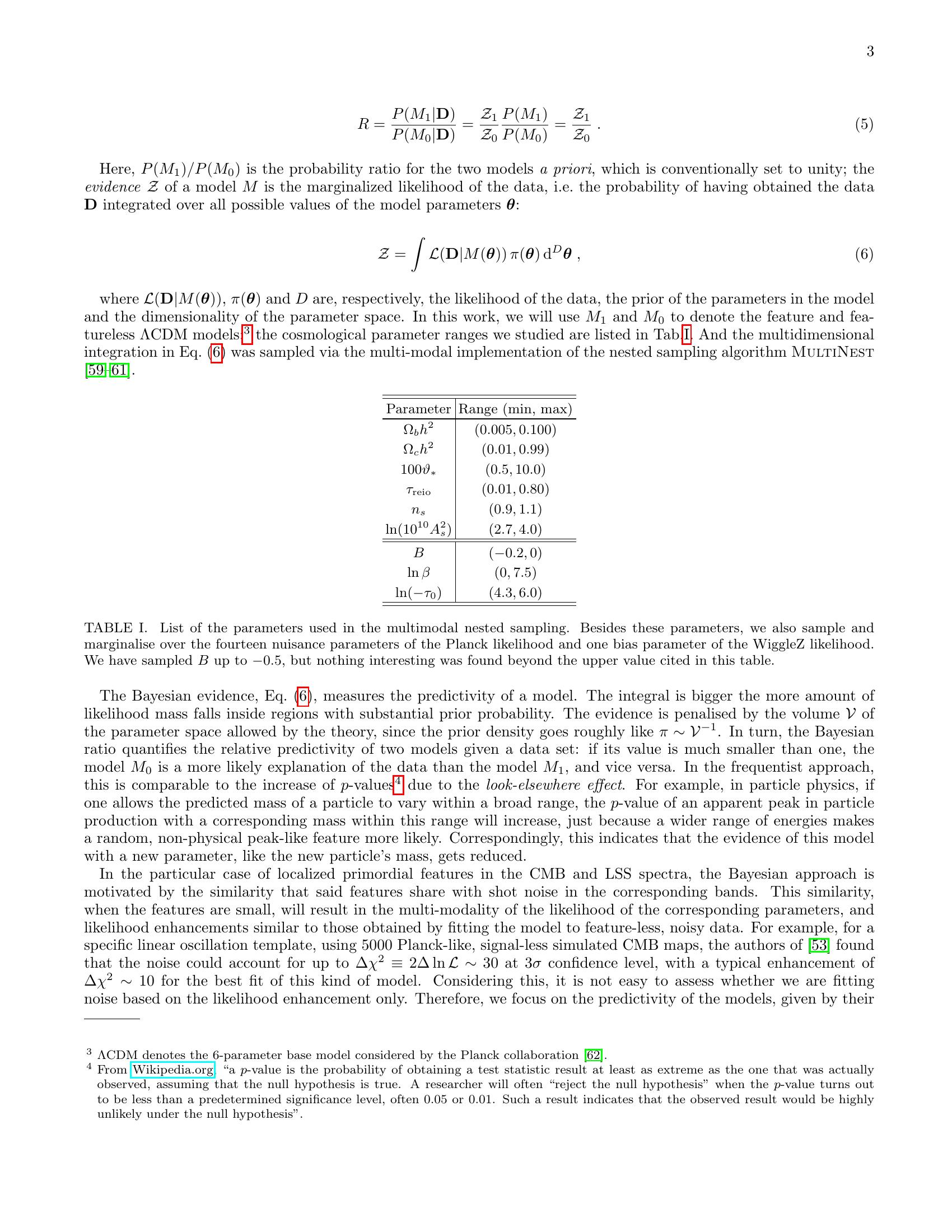}
    \end{subfigure}
    \hfill
    \begin{subfigure}[t]{0.40\textwidth}
        \includegraphics[max height=7cm]{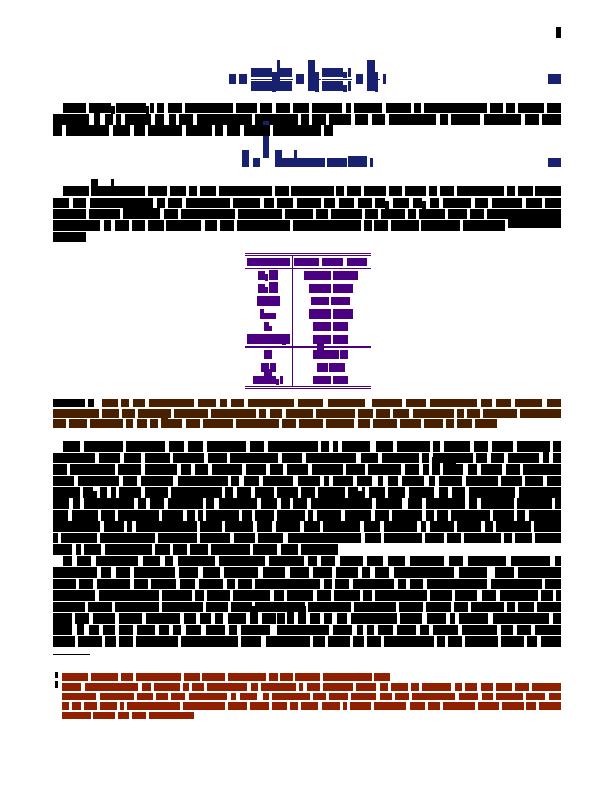}
    \end{subfigure}
    \caption{DocBank sample image on the left and its corresponding segmentation on the right. Each color represents one class (black for \textit{paragraph}, purple for \textit{equation}, ...).}
    \label{fig:docbank_sample}
\end{figure}
\vspace{-5mm}

\section{Models}

We compared LayoutLM, a Transformer for DU which is our baseline, with our long range contributions LayoutLinformer and LayoutCosformer\footnote{Models implementation and weights available at \url{https://github.com/thibaultdouzon/long-range-document-transformer}}.
They only differ by their implementation of self-attention: LayoutLM uses full self-attention like BERT, LayoutLinformer uses a low-rank approximation first proposed by~\cite{Wang2020} and LayoutCosformer uses a kernel-based method introduced in~\cite{Qin2022} as a replacement.
We further detail how they work in the subsequent subsections.

We chose those models over other efficient Transformers based on the convenience to adapt them from linear text to 2-dimensional documents.
Efficient attention based on sliding windows~\cite{Beltagy2020, Zaheer2020} does not transpose nicely to 2D documents because the sliding window mechanism is deeply linked to the linear order of words.
Even though our approach tries to provide words in a natural order, in some documents it does not reflect the human reading order -- for example for table content.
To mitigate this issue, we preferred to rely on global attention or 2D local attention.


Similarly to how LayoutLM was adapted from BERT, we adapt Linformer and cosFormer models to process documents by adding a 2D positional embedding and a page embedding to the input.
We chose to use learned embeddings to simplify weight transfer from LayoutLM to our long-range models.

\subsection{LayoutLM}
\begin{figure}[htp]
    \centering
    \includegraphics[scale=0.5]{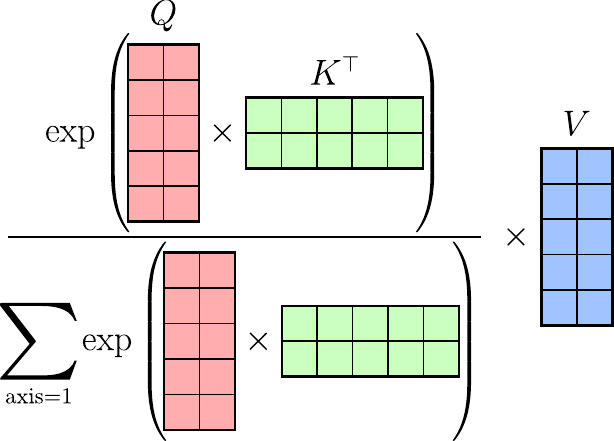}
    \caption{Illustration of the attention mechanism used in LayoutLM, normalization and multiple heads aside. In this example, $N = 5$ and $d = 2$. Due to the softmax operator, the product $QK^\top$ must be computed, resulting in $O(N^2)$ complexity.}
    \label{fig:transformer_equation}
\end{figure}

LayoutLM~\cite{Xu2020} has proven its capacities on most tasks related to documents since its release. 
It reuses BERT~\cite{Devlin2019} encoder and tokenizer, and only modifies the positional encoding by introducing a 2D encoding for word boxes boundary and size.
This modification allows the model to leverage layout information provided by the OCR.
LayoutLM's computational bottleneck is the self-attention layer.
In Transformers, self-attention~\cite{Vaswani2017} takes \textit{queries} $Q$, \textit{keys} $K$ and \textit{values} $V$ and computes a weighted average of \textit{values} for each input.
The weights are given by the dot product between each pair of \textit{queries} and \textit{keys}.
It can be formulated $\texttt{softmax}(QK^\top)V$, where $Q, K, V \in \mathbb{R}^{N\times d}$ and $N$ represents the sequence length and $d$ the model hidden size.
\figurename~\ref{fig:transformer_equation} describes the self-attention operation.
The matrix $\texttt{softmax}(QK^\top)$ is the attention matrix containing the intensity of attention between each pair of tokens.
\citet{Xu2020} pre-trained the model on RVL-CDIP~\cite{Harley2015} which contains 7 millions scanned documents released in the 90' from the tobacco industry.
Two versions of LayoutLM was have been released: base and large, and it outperforms all preceding text-only language models on classification and information retrieval tasks.

\vspace{-5mm}
\vspace{-5mm}

In our experiments, we only use the base model with maximum sequence length $N = 512$ and hidden size $d = 768$.
For longer documents, we split the tokenized sequence into chunks of maximum length and process them separately.

\subsection{LayoutLinformer}
\begin{figure}[htp]
    \centering
    \includegraphics[scale=0.5]{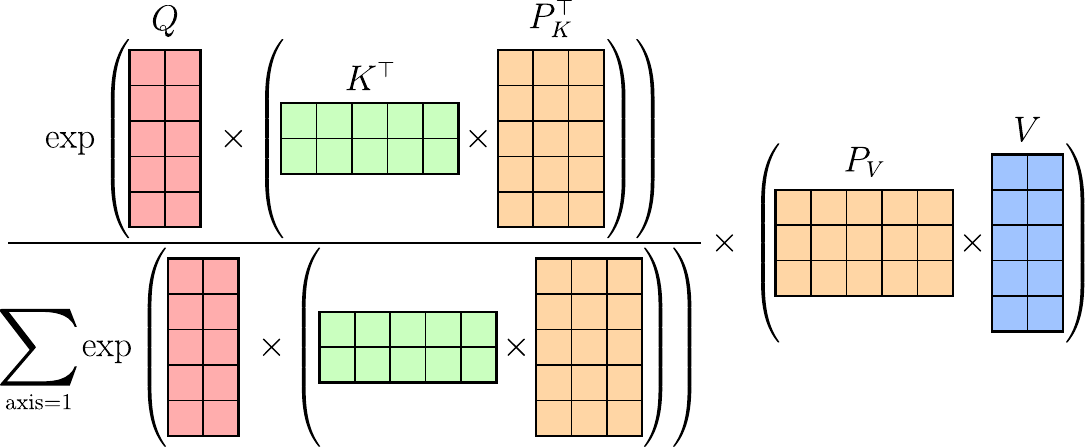}
    \caption{LayoutLinformer attention mechanism. In this example, $N = 5$, $d = 2$ and $k = 3$. Efficient matrix multiplication ordering reduces the complexity to $O(Nk)$.}
    \label{fig:linformer_equation}
\end{figure}

Our first contribution, LayoutLinformer is based on the Linformer architecture~\cite{Wang2020} and adapted to document processing by adding 2D positional encodings and using 
 LayoutLM pre-trained weights.
Although true self-attention can only be computed in $O(N^2)$, it can be approximated very efficiently by leveraging the low rank of the attention matrix $QK^\top$.
In \figurename~\ref{fig:linformer_equation}, we illustrate LayoutLinformer's attention mechanism.
Keys and values sequence length dimension is projected on a smaller space of size $k$ through a linear transformation: $K' = P_KK$ where $P_k \in \mathbb{R}^{k\times N}$ is the learned projection matrix (respectively $V' = P_VV$ where $P_V \in \mathbb{R}^{k\times N}$).
This means the size of the new attention matrix $Q(P_KK)^\top$ is $N \times k$, reducing the complexity of self-attention to $O(Nk)$.

\vspace{-5mm}
\vspace{-5mm}

An immediate drawback of this projection is the loss of ability to visualize the attention matrix in order to explain the model.
It is also no longer possible to implement causal attention or any specific attention pattern.
On the other hand, Linformer provides a simple modification to the Transformer in order to make it manage longer sequences with global attention.
Most model weights are identical between the two architectures, allowing us to transfer LayoutLM pre-trained weights into DocumentLinformer before further pre-training.

\citet{Wang2020} showed that it can obtain a performance comparable to Roberta~\cite{Liu2019} on multiple NLP benchmarks.
He brang evidence that its performance is mostly determined by the projection dimension $k$, and that increasing sequence length $N$ did not degrade results.
Therefore, we chose to apply LayoutLinformer with $N = 2048$ and $k = 512$ in order to compare its performances with LayoutLM.

\subsection{LayoutCosformer}

\begin{figure}[htp]
    \centering
    \includegraphics[scale=0.5]{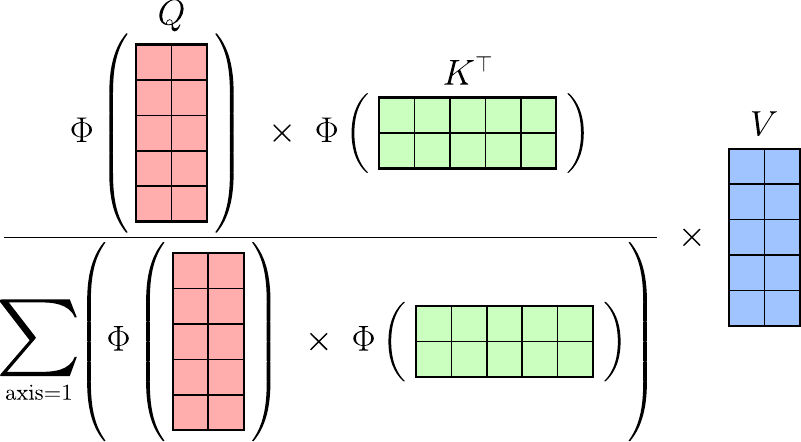}
    \caption{LayoutCosformer efficient attention mechanism with $N = 5$ and $d = 2$. The linear similarity enable computing first $\Phi(K^\top)V$ and factorize $\Phi(Q)$ out of the summation.}
    \label{fig:cosformer_equation}
\vspace{-5mm}
\end{figure}

Our second contribution, called LayoutCosformer, is based on the cosFormer~\cite{Qin2022} model which is another efficient alternative to the original Transformer.
Similarly to LayoutLinformer, we transferred pre-trained weights from LayoutLM to DocumentCosFormer thanks to the similarities between architectures.
It achieves linear complexity by replacing the non-linear similarity computation between $Q$ and $K$ with a linear operation.
More specifically, \citet{Qin2022} proposed to replace $\exp(QK^\top)$ with $\Phi(Q)\Phi(K^\top)$ where $\Phi$ is a nonlinear function.
\figurename~\ref{fig:cosformer_equation} illustrates in more detail how LayoutCosformer attention works.
In order to keep values of the similarity matrix positive, a good choice is $\Phi = \texttt{ReLU}$.
Computations can then be reordered to decrease the complexity to $O(N)$.

In addition to its linear self-attention complexity, \citet{Qin2022} include a relative self-attention bias towards nearby tokens. 
They cannot simply add the bias to the $N \times N$ similarity matrix before multiplying with values because it would mean a quadratic complexity.
Their solution is to use functions that can be decomposed into a sum of products: $f(x, y) = \sum_n g_n(x)\times h_n(y)$.
If we call $B$ the bias matrix where $B_{i, j} = f(i, j)$, their biased similarity matrix can be written $\Phi(Q)\Phi(K^\top)\odot B$ where $\odot$ is the element-wise product.
Then when looking at the attention from token $i$ to token $j$ we obtain:

\vspace{-1mm}
$$
\begin{aligned}
s_{i, j} &= \Phi(Q_i)\Phi(K_j^\top) B_{i, j}\\
&= \Phi(Q_i)\Phi(K_j^\top) \sum_n g_n(i)\times h_n(j)\\
&= \sum_n \Phi(Q_i)\Phi(K_j^\top) g_n(i) h_n(j)\\
&= \sum_n (\Phi(Q_i) g_n(i)) \times (\Phi(K_j^\top)h_n(j))
\end{aligned}
$$
\vspace{-3mm}

Using this trick, they proposed to use a cosine bias $B_{i, j} = \cos(\frac{\pi}{2M} (i-j))$ which can be decomposed into $B_{i, j} = \cos(\frac{\pi}{2M}i)\cos(\frac{\pi}{2M}j) + \sin(\frac{\pi}{2M}i)\sin(\frac{\pi}{2M}j)$. 
With the normalization constant $M$ set to the maximum sequence length, they ensure $0 < B_{i, j} < 1$ with a maximum when $i=j$.
In the next subsection, we demonstrate how it can also be applied to 2D relative attention.

\subsection{2D Relative attention}

Global self-attention is a powerful tool for capturing long-range dependencies. 
However, although distant dependencies can be relevant, most attention should be toward close neighbors.
Relative attention~\cite{Shaw2018, Powalski2021} selectively focuses on specific parts of the input by biasing the base self-attention.
This was proven useful on text which that can be represented as a linear sequence, but due to complex layouts, the sequence order is suboptimal to determine locality.
In order to better capture local context in documents, we introduced 2D relative attention based on the token positions inside the document.

In LayoutLM, we pre-compute for each document an attention bias matrix $B$ and modify the self-attention formula to take it into account.
More precisely, we replace the self-attention with:

\vspace{-1mm}
$$
\mathrm{RelativeAttention}(Q, K, V, B) = \left(\mathrm{softmax}(QK^\top)\odot B\right)V
$$
\vspace{-3mm}

Where $\odot$ denotes element-wise multiplication.
Directly multiplying the attention matrix by some bias is very flexible and allows for any bias matrix to be chosen.
It also matches the way LayoutCosformer applies relative bias to its self-attention, thus allowing to compare them.

On the other hand, it is nontrivial to implement relative attention for global long-range Transformers.
Because LayoutLinformer compresses the sequence dimension of the Key matrix, it is not possible to apply custom 2D attention bias to 
LayoutLinformer.
For LayoutCosformer it is possible to reuse the same trick as in the 1D version with another bias function.

Because the function must remain separable into a sum of products, a good choice is to use exponentials and trigonometric functions.
We first prove that the product of two separable functions is also itself separable.
Let $f^1 = \sum_n g^1_n(x) \times h^1_n(y)$ and $f^2 = \sum_m g^2_m(x) \times h^2_m(y)$ be two functions separable into sum of products, then:

\vspace{-0.6cm}
$$
\begin{aligned}
f^1(x, y) \times f^2(x, y) &= \left(\sum_n g^1_n(x) \times h^1_n(y)\right) \times \left(\sum_m g^2_m(x) \times h^2_m(y)\right) \\
&= \sum_n \sum_m \left(g^1_n(x) \times h^1_n(y) \times g^2_m(x) \times h^2_m(y) \right) \\
&= \sum_{n, m} (g^1_n(x) g^2_m(x)) \times (h^1_n(y) h^2_m(y))
\end{aligned}
$$

Which can also be separated into a sum of products.

We chose to compare 2 different attention biases.
The first one is simply the product cosine bias along both X and Y axis.
It captures local context in every direction with variations close to euclidean distance.
We define $B^{\mathrm{squircle}}$ \footnote{Squircle are intermediate shape between square and circle, see \url{https://en.wikipedia.org/wiki/Squircle}. Contours of the surface described by $B^{\mathrm{squircle}}$ is not actually a squircle but also range from square to circle.} the following:

\vspace{-1mm}
$$
B^{\mathrm{squircle}}_{i, j} = \cos(\frac{\pi}{2M}(x_i-x_j)) \times \cos(\frac{\pi}{2M}(y_i-y_j))
$$
\vspace{-1mm}

Where $x_i$ and $y_i$ (resp. $x_j$ and $y_j$) are positions of token $i$  (resp. $j$) along X and Y axis.
In practice we used the coordinates of the center of each token bounding box.

\begin{figure}[t]
\centering
    \begin{subfigure}[t]{0.46\textwidth}
    \includegraphics[max width=\textwidth, max height=3cm]{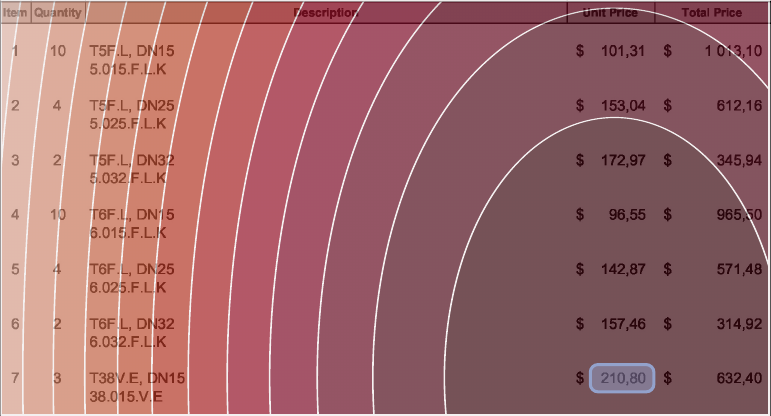}
    \caption{Squircle relative attention bias.}
  \end{subfigure}\hfill
  \begin{subfigure}[t]{0.46\textwidth}
      \includegraphics[max width=\textwidth, max height=3cm]{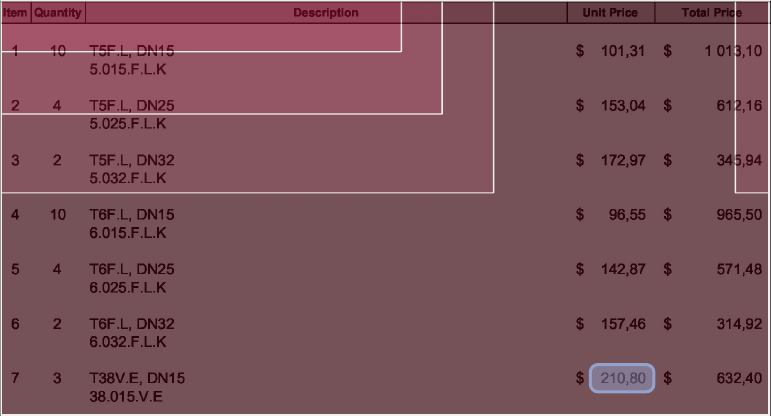}
      \caption{Cross relative attention bias.}
  \end{subfigure}
  \begin{subfigure}[t]{0.05\textwidth}
      \includegraphics[max width=\textwidth, max height=3cm, trim=0 0 10px 3px]{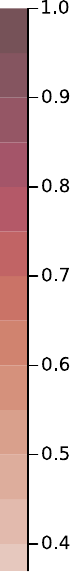}
  \end{subfigure}
    \caption{Contour plots for squircle and cross relative attention bias applied to token ``210,80'' (bottom-right corner). Because token positions are normalized between 0 and 1000, tokens along the same line cannot fully attend to each other on the left while they are unaffected on the right.}
    \label{fig:tableau_po}
    \vspace{-5mm}
\end{figure}

Although this bias correctly captures 2D locality, documents complex layout sometimes implicitly calls for other definition of proximity in order to understand it.
For instance, \figurename~\ref{fig:tableau_po} shows a table from a purchase order. 

In this configuration, in order to grasp correctly the meaning of a cell in the table, the model needs to make the connection with the table header positioned at the beginning of the page. 
When multiple line items are spanning the whole page, we hypothesize that this relative attention might hurt the performance due to the long-distance separating tokens.
To deal with this issue, we propose another bias pattern.
Its objective is to allow attention to tokens that are aligned with each other along the X or Y axis.
To this end, we define $B^{\mathrm{cross}}_{i, j} = \max\{\cos(\frac{\pi}{2M}(x_i-x_j)) , \cos(\frac{\pi}{2M}(y_i-y_j))\}$.
We illustrate the differences with an example shown in \figurename~\ref{fig:tableau_po}. 
With cross relative attention bias, 
the highlighted token (the price of an item) can better attend to the column header ``Unit Price'' and to its related line.
In general, tokens inside a table can fully attend to their corresponding column header and line.
This should prove helpful for understanding tables by guiding the model attention towards semantically related tokens.

\vspace{-2mm}
\section{Experiments}

Our models are pre-trained on our Business Documents collection for 200k steps using Masked Visual-Language Modeling~\cite{Xu2020}.
They are then finetuned on each dataset.
For both tasks, we use \texttt{BIESO} tags to help the model decode predictions spanning multiple tokens.
We performed our experiments on two RTX A6000 for pre-trainings and single RTX A6000 for fine tunings. 
LayoutLM models runs with a batch size of 48 and sequence length of 512 while long-range models (LayoutLinformer and LayoutCosFormer) can only get to a batch size of 16 with sequence length of 2048 on a single device.
We accumulate gradient for 96 data samples before updating model's weights.
We use Adam with learning rate $lr = 2 \cdot 10^{-5}$ and linear warmup for 5\% of the training steps followed by linear decrease.

\vspace{-2mm}
\subsection{Long-Range}

\begin{table}[h!]
    \centering
    \addtolength{\leftskip} {-1.5cm}
    \addtolength{\rightskip}{1.5cm}
    \begin{tabular}{c|R{0.8cm} C{0.1cm} L{0.9cm} R{0.8cm} C{0.1cm} L{0.9cm} R{0.8cm} C{0.1cm} L{0.9cm} R{0.8cm} C{0.1cm} L{0.9cm} R{0.8cm} C{0.1cm} L{0.9cm} R{0.8cm} C{0.1cm} L{0.9cm}}
         & \multicolumn{18}{c}{Time in seconds / \emph{Memory in GiB}} \\
         & \multicolumn{18}{c}{Sequence length} \\
         Model name & \multicolumn{3}{c}{512} & \multicolumn{3}{c}{1024} & \multicolumn{3}{c}{2048} & \multicolumn{3}{c}{4096} & \multicolumn{3}{c}{8192} & \multicolumn{3}{c}{16384} \\
         \hline
         LayoutLM & 1.41 & & $ \emph{1.25} $ & 2.83 & & $ \emph{2.50} $ & 7.39 & & $ \emph{5.01} $ & 23.43 & & $ \emph{13.69} $ & \multicolumn{3}{c}{-} & \multicolumn{3}{c}{-} \\
         LayoutLinformer & 1.18 & & $ \emph{1.35} $ & 1.92 & & $ \emph{2.26} $ & 3.54 & & $ \emph{3.28} $ & 6.90 & & $ \emph{5.19} $ & 13.08 & & $ \emph{8.96} $ & 25.65 & & $ \emph{16.78} $\\
         LayoutCosformer & 2.03 & & $ \emph{1.36} $ & 2.50 & & $ \emph{2.37} $ & 4.68 & & $ \emph{3.38} $ & 9.00 & & $ \emph{5.38} $ & 17.23 & & $ \emph{9.59} $ & 33.96 & & $ \emph{17.59} $
    \end{tabular}
    \caption{Duration and memory consumption of the 3 models for various sequence lengths on an inference task.}
    \label{tab:efficiency}
    \vspace{-8mm}
\end{table}

Theoretical results on models architectures hints towards LayoutLinformer and LayoutCosformer being much more efficient the longer the sequence.
We use a dummy inference task with increasing sequence lengths and compare our 2 models with LayoutLM base architecture.
The results are available in table~\ref{tab:efficiency}.
They reveal how the computational complexity of full self-attention disables LayoutLM when dealing with sequence longer than 1024. 
Its memory consumption limits our tests with LayoutLM up to sequence length of 4096, longer sequences couldn't fit into a single GPU.
On the other hand, LayoutLinformer and LayoutCosformer performed as predicted, with LayoutCosformer being slightly slower and more memory hungry than LayoutLinformer.

It turns out document's length also greatly impacts models metrics performance on the Customer Order dataset.
For better visualization, we group documents into 3 length categories: short (document fits into 512 tokens), medium (between 513 and 2048) and long (2049 or more tokens).
LayoutLM models can process short documents in a single sequence but need to split other documents into multiple independent sequences.
Short and medium documents fit into LayoutLinformer and LayoutCosformer sequence length but not long documents.
When a model cannot process a document in a single sequence, we split the document into multiple sequences and process them separately.


\begin{figure}[h!]
    \centering
    \addtolength{\leftskip}{-2cm}
    \addtolength{\rightskip}{-2cm}
    \begin{subfigure}[t]{0.5\textwidth}
        \includegraphics[max width=\textwidth, trim=4cm 0.5cm 6cm 1cm]{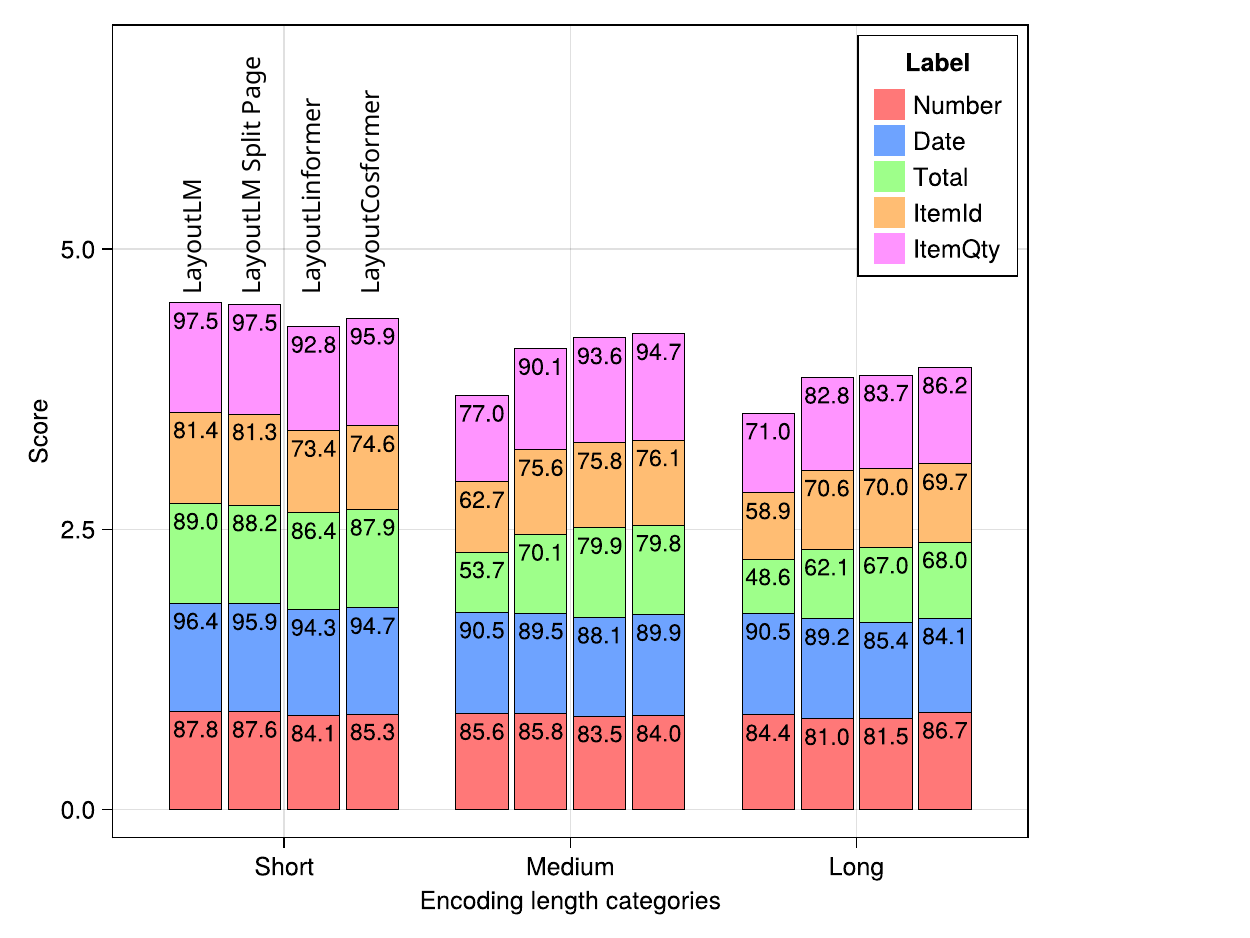}
    \end{subfigure}
    \caption{F1-score stacked bar plot of multiple models on the Business Orders dataset. In each document length categories, models are in the same order.}
    \label{fig:plot_order_base}
\vspace{-5mm}
\end{figure}


In \figurename~\ref{fig:plot_order_base}, we compare our pre-trained LayoutLM models with LayoutLinformer and LayoutCosformer.
First, we discovered LayoutLM is very sensitive to the split position for medium and long documents.
Introducing a sequence split when a new page is started greatly improves performance, we call this model LayoutLM SplitPage.
It performs better on \textit{total amount} (from 53.7\% to 70\%), item ID number (from 62.7\% to 75.6\%) and \textit{quantity} (from 77.0\% to 90.1\%) recognition for medium and long documents.
The repetitive structure of multipage documents combined with the fact that most pages fit in a 512 tokens sequence allow the model to not get lost.
\textit{Document number} and \textit{date} are mostly not affected because they almost always occur at the beginning of the document, which is not affected by the splitting strategy.

Although LayoutLinformer and LayoutCosformer perform slightly worse than LayoutLM for short documents on all classes (around 74\% F1 score on \textit{item ID number} versus 81\% for LayoutLMs), their performance decreases less than LayoutLM's on medium documents.
On those medium documents, even LayoutLM SplitPage drops from 88.2\% to 70.1\% F1 score on the \textit{total amount} while both long-range models only reduce performance from roughly 87\% to 80\%.
We also noticed date recognition performance degrades across all models with longer documents which is not expected because \textit{dates} are usually at the top of the first page.
The same can be noted for the \textit{order number} at a smaller scale.
It might be due to a correlation between document's length and layout: short and medium~/~long documents do not share layouts. 
And because there are twice more short documents than longer ones, it is harder to generalize to new layouts.
Overall, the performance of long-range models is more consistent acrossa wide variety of document lengths.

\vspace{-7mm}
\begin{table}[htp]
    \centering
    \begin{tabular}{c|L{1.5cm} C{1.5cm} R{1.5cm}}
         & \multicolumn{3}{c}{F1 weighted macro average} \\
        Model name & Short & Medium & Long \\
        \hline
         LayoutLM                           & 95.36 & 95.84 & 91.42 \\
         LayoutLinformer                    & 95.20 & 96.49 & 91.41 \\
         LayoutCosformer                    & 94.03 & 95.91 & 91.40 \\
    \end{tabular}
    \caption{F1 weighted average for each model and document length categories. All models were first pre-trained on the Business Documents collection.}
    \label{tab:docbank_cat_len}
\end{table}
\vspace{-9mm}

We performed the same experiments on Docbank dataset, except for the page-splitting part as all documents are single page.
At first we compared models performance for each document length categories in \tableautorefname~\ref{tab:docbank_cat_len}.
It contains average F1 score across all labels weighted by the support of each label.
It turns out length categories introduce bias in the composition of pages, with labels being very sparsely represented in some categories.
This bias implicitly selects more first page in short pages (with lower text density), and medium sized pages contain a lot of paragraphs.

We observe the same drop in performance for long-range models on short documents, with LayoutLinformer providing better results across the board than LayoutCosformer.
But we notice LayoutLM perform slightly better on medium documents than on short.
Long-range models follow the same pattern with a greater difference between short and medium pages, LayoutCosformer almost gaining 2 average F1 percentage points.
There is almost 20 times fewer long documents than medium, which could explain part of the global performance loss.
Unfortunately, due to those biases, it is difficult to draw conclusions on model's performance.

\begin{table}[htp]
\vspace{-4mm}
    \centering
    \addtolength{\leftskip} {-3cm}
    \addtolength{\rightskip}{2.5cm}
    \begin{tabular}{c c|c c c c c c c c c c c c|c}
     & 2D Relative & \multicolumn{12}{c|}{F1 score} & \multicolumn{1}{c}{Macro} \\
    Model name & attention & Abst. & Author & Caption & Equa. & Figure & Footer & List & Para. & Refe. & Section & Table & Title & average\\
    \hline
    LayoutLM (\citet{Li2020}) & - &  98.1 &  85.9 &  95.9 &  89.4 & \textbf{100.0} & 89.5 & \textbf{89.4} &  \textbf{97.8} &  93.3 &  \textbf{95.9} &  86.3 &  \textbf{95.7} &  93.1 \\
    LayoutLM (\citet{Xu2020}) & - & 98.3 & 89.6 & 96.0 & 89.0 & 99.7 & 91.6 & 88.2 & 97.5 & \textbf{93.5} & 94.3 & 87.4 & 90.4 & 93.0\\
    LayoutLM (*) & - & 97.8 & 87.5 & 94.9 & 87.2 & 99.7 & 90.5 & 84.0 & 97.1 & 93.2 & 92.8 & 85.7 & 88.6 & 91.6\\
    LayoutLM (*) & Squircle  & \textbf{98.4} & 90.2 & \textbf{96.1} & 89.7 & 99.8 & 92.0 & 88.9 & 97.6 & 93.4 & 94.6 & 87.7 & 90.3 & \textbf{93.2}\\
    LayoutLM (*) & Cross & \textbf{98.4} & \textbf{90.3} & 96.0 & 89.6 & 99.8 & \textbf{92.1} & 88.7 & 97.6 & 93.4 & 94.6 & 87.5 & 90.7 & \textbf{93.2}\\
    & & & & & & & & & & & & &\\
    LayoutLinformer (*) & - & 97.9 & 88.9 & 93.7 & 90.0 & 99.5 & 91.1 & 87.9 & 97.5 & 93.2 & 91.3 & 87.6 & 88.7 & 92.3\\
    LayoutCosformer (*) & - & 97.2 & 87.2 & 91.0 & 88.1 & 99.3 & 90.6 & 87.4 & 97.1 & 93.2 & 81.4 & 87.0 & 88.3 & 90.7\\
    LayoutCosformer (*) & Squircle & 97.0 & 85.4 & 92.4 & 89.2 & 98.8 & 90.7 & 84.2 & 97.2 & 93.2 & 85.6 & 87.9 & 86.8 & 90.7\\
    LayoutCosformer (*) & Cross & 97.4 & 86.9 & 93.8 & \textbf{91.2} & 98.9 & 91.7 & 87.5 & 97.5 & 93.1 & 87.4 & \textbf{89.0} & 88.1 & 91.9\\
    \end{tabular}
    \caption{Results on Docbank dataset for LayoutLMs and long-range models. Models with a asterisk (*) are ours. They were pre-trained on the Business Documents collection before finetuning on Docbank.}
    \label{tab:docbank_base_results}
    
\vspace{-8mm}
\end{table}

\tableautorefname~\ref{tab:docbank_base_results} compiles results for LayoutLM and long-range models for all labels.
First, we can make sure our training pipeline performs on par with what Docbank authors reported for LayoutLM base model by comparing their results and the ones we obtained by using public LayoutLM weights.
Except for \textit{author} and \textit{title} labels, both results are very close, and the macro average is almost identical.
Secondly, pre-training on business documents negatively impacts LayoutLM performances on all labels, losing 1.4 F1 percentage points on average.
This advocates for pre-training data crucial role in later model finetuning results and its composition.
Finally, long-range models performed on the same level as LayoutLM.
LayoutLinformer even being more performant than our pre-trained LayoutLM.
Overall, even though LayoutCosformer seems less performant on this task, both long-range models performed better than our pre-trained LayoutLM on \textit{table} and \textit{equation}.
Those two labels might beneficiate from long-range references, giving the model hints of their presence in the current sequence.

\vspace{-2mm}
\subsection{Relative Attention}
We conduct the same experiments on models with 2D relative attention and compare their performance with their flat attention counterpart.
On the business order dataset, \tableautorefname~\ref{tab:relative_attention} shows slight gains when using squircle attention with LayoutLM.
For all document lengths, information retrieval is improved a few percentage points of F1 score over our previous LayoutLM Split Page implementation.
Though, we do not observe the same improvement with the cross shaped attention pattern.
This might indicate focusing on very local neighbors helps LayoutLM making the right decision.
Overall, relative attention improves results in some circumstances but not as much as splitting every page did.
However, when combined with LayoutCosformer, we observe a significant degradation in performance for all labels with the squircle attention while the cross pattern provides similar results as the raw LayoutCosformer.


\begin{table}[h!]
\vspace{-4mm}
    \centering
    \begin{tabular}{c c|c c c}
        & & \multicolumn{3}{c}{Macro average F1 score}\\
        Model name & 2D Relative attention & Short & Medium & Long \\
        \hline
        LayoutLM Split Page & - & \num[round-mode=places,round-precision=1]{\fpeval{round(450 / 5, 1)}} & \num[round-mode=places,round-precision=1]{\fpeval{round(411 / 5, 1)}} & \num[round-mode=places,round-precision=1]{\fpeval{round(386 / 5, 1)}} \\
        LayoutLM Split Page & Squircle & \textbf{\num[round-mode=places,round-precision=1]{\fpeval{round(452 / 5, 1)}}} & \num[round-mode=places,round-precision=1]{\fpeval{round(415 / 5, 1)}} & \num[round-mode=places,round-precision=1]{\fpeval{round(389 / 5, 1)}} \\
        LayoutLM Split Page & Cross & \num[round-mode=places,round-precision=1]{\fpeval{round(450 / 5, 1)}} & \num[round-mode=places,round-precision=1]{\fpeval{round(410 / 5, 1)}} & \num[round-mode=places,round-precision=1]{\fpeval{round(388 / 5, 1)}} \\
        & & & & \\
        LayoutCosformer & - & \num[round-mode=places,round-precision=1]{\fpeval{round(438 / 5, 1)}} & \num[round-mode=places,round-precision=1]{\fpeval{round(425 / 5, 1)}} & \textbf{\num[round-mode=places,round-precision=1]{\fpeval{round(395 / 5, 1)}}} \\
        LayoutCosformer & Squircle & \num[round-mode=places,round-precision=1]{\fpeval{round(429 / 5, 1)}} & \num[round-mode=places,round-precision=1]{\fpeval{round(411 / 5, 1)}} & \num[round-mode=places,round-precision=1]{\fpeval{round(367 / 5, 1)}} \\
        LayoutCosformer & Cross & \num[round-mode=places,round-precision=1]{\fpeval{round(438 / 5, 1)}} & \textbf{\num[round-mode=places,round-precision=1]{\fpeval{round(426 / 5, 1)}}} & \num[round-mode=places,round-precision=1]{\fpeval{round(386 / 5, 1)}}
        
    \end{tabular}
    \caption{Macro average F1 score on the Business Orders dataset with 2D relative attention.}
    \label{tab:relative_attention}
    \vspace{-8mm}
\end{table}

On Docbank task, relative attention provides noticeable performance gains for both LayoutLM and LayoutCosformer.
We provide all results in \tableautorefname~\ref{tab:docbank_base_results}.
LayoutLM with relative attention is standing out, going from 91.6\% F1 score to 93.2\% for both squircle and cross patterns.
Most improvements are made on \textit{author}, \textit{equation} and \textit{list}, each gaining at least 2 F1 score points.
Both resulting models even beat Docbank's authors version by a thin margin.
This is impressive knowing those models were pre-trained on the same business order dataset as our base LayoutLM which suffered a 1.5 F1 score performance drop as a consequence.
It turns out \textit{author}, \textit{equation} and \textit{list} were also the fields where our LayoutLM performance dropped the most compare to stock LayoutLM.
Applying cross shaped relative attention to LayoutCosformer also improves performance across most labels.
It even outperforms all other models on \textit{equation} and \textit{table} fields which benefit most from very long attention.

\vspace{-2mm}
\section{Conclusion}
\vspace{-1mm}

In this work, we showed the impact of document length on Transformer-based models applied to Document Understanding.
Depending on the document's type and the task, model's performance on longer documents can be negatively impacted with F1 score dropping 20\% for the most impacted.
We explored several alternatives including another sequence split strategy and long-range layout-aware models based on Linformer and cosFormer architectures.
They all proved to successfully reduce the performance gap between short and long documents (down to only 10\% performance drop), sometimes at a small cost on short document's metrics.
We also introduce relative attention based on 2D textual layout instead of the classical sequence order.
It produces better results on dense text, significantly improving both LayoutLM and LayoutCosformer on the Docbank layout segmentation task.

In addition to other efficient Transformer architectures, we plan to investigate other ways to use longer sequences for DU.
For example, in multi-modal models, this may allow fitting the whole text and visual patches of a document in a single sequence without needing more compute capabilities.

\bibliographystyle{plainnat}
\bibliography{icdar_2023}

\begin{thebibliography}{34}
\providecommand{\natexlab}[1]{#1}
\providecommand{\url}[1]{\texttt{#1}}
\expandafter\ifx\csname urlstyle\endcsname\relax
  \providecommand{\doi}[1]{doi: #1}\else
  \providecommand{\doi}{doi: \begingroup \urlstyle{rm}\Url}\fi

\bibitem[Ainslie et~al.(2020)Ainslie, Ontanon, Alberti, Cvicek, Fisher, Pham, Ravula, Sanghai, Wang, and Yang]{Ainslie2020}
Joshua Ainslie, Santiago Ontanon, Chris Alberti, Vaclav Cvicek, Zachary Fisher, Philip Pham, Anirudh Ravula, Sumit Sanghai, Qifan Wang, and Li~Yang.
\newblock {ETC}: Encoding long and structured inputs in transformers.
\newblock In \emph{Proceedings of the 2020 Conference on Empirical Methods in Natural Language Processing ({EMNLP})}, pages 268--284, Online, November 2020. Association for Computational Linguistics.
\newblock \doi{10.18653/v1/2020.emnlp-main.19}.
\newblock URL \url{https://aclanthology.org/2020.emnlp-main.19}.

\bibitem[Bahdanau et~al.(2016)Bahdanau, Cho, and Bengio]{Bahdanau2016}
Dzmitry Bahdanau, Kyunghyun Cho, and Yoshua Bengio.
\newblock Neural machine translation by jointly learning to align and translate, May 2016.
\newblock URL \url{http://arxiv.org/abs/1409.0473}.
\newblock type: article.

\bibitem[Beltagy et~al.(2020)Beltagy, Peters, and Cohan]{Beltagy2020}
Iz~Beltagy, Matthew~E. Peters, and Arman Cohan.
\newblock Longformer: The long-document transformer, December 2020.
\newblock URL \url{http://arxiv.org/abs/2004.05150}.
\newblock type: article.

\bibitem[Brown et~al.(2020)Brown, Mann, Ryder, Subbiah, Kaplan, Dhariwal, Neelakantan, Shyam, Sastry, Askell, Agarwal, Herbert-Voss, Krueger, Henighan, Child, Ramesh, Ziegler, Wu, Winter, Hesse, Chen, Sigler, Litwin, Gray, Chess, Clark, Berner, {McCandlish}, Radford, Sutskever, and Amodei]{Brown2020}
Tom Brown, Benjamin Mann, Nick Ryder, Melanie Subbiah, Jared~D Kaplan, Prafulla Dhariwal, Arvind Neelakantan, Pranav Shyam, Girish Sastry, Amanda Askell, Sandhini Agarwal, Ariel Herbert-Voss, Gretchen Krueger, Tom Henighan, Rewon Child, Aditya Ramesh, Daniel Ziegler, Jeffrey Wu, Clemens Winter, Chris Hesse, Mark Chen, Eric Sigler, Mateusz Litwin, Scott Gray, Benjamin Chess, Jack Clark, Christopher Berner, Sam {McCandlish}, Alec Radford, Ilya Sutskever, and Dario Amodei.
\newblock Language models are few-shot learners.
\newblock In \emph{Advances in Neural Information Processing Systems}, volume~33, pages 1877--1901. Curran Associates, Inc., 2020.
\newblock URL \url{https://papers.nips.cc/paper/2020/hash/1457c0d6bfcb4967418bfb8ac142f64a-Abstract.html}.

\bibitem[Chowdhery et~al.(2022)Chowdhery, Narang, Devlin, Bosma, Mishra, Roberts, Barham, Chung, Sutton, Gehrmann, Schuh, Shi, Tsvyashchenko, Maynez, Rao, Barnes, Tay, Shazeer, Prabhakaran, Reif, Du, Hutchinson, Pope, Bradbury, Austin, Isard, {Gur-Ari}, Yin, Duke, Levskaya, Ghemawat, Dev, Michalewski, Garcia, Misra, Robinson, Fedus, Zhou, Ippolito, Luan, Lim, Zoph, Spiridonov, Sepassi, Dohan, Agrawal, Omernick, Dai, Pillai, Pellat, Lewkowycz, Moreira, Child, Polozov, Lee, Zhou, Wang, Saeta, Diaz, Firat, Catasta, Wei, {Meier-Hellstern}, Eck, Dean, Petrov, and Fiedel]{chowdheryPaLMScalingLanguage2022}
Aakanksha Chowdhery, Sharan Narang, Jacob Devlin, Maarten Bosma, Gaurav Mishra, Adam Roberts, Paul Barham, Hyung~Won Chung, Charles Sutton, Sebastian Gehrmann, Parker Schuh, Kensen Shi, Sasha Tsvyashchenko, Joshua Maynez, Abhishek Rao, Parker Barnes, Yi~Tay, Noam Shazeer, Vinodkumar Prabhakaran, Emily Reif, Nan Du, Ben Hutchinson, Reiner Pope, James Bradbury, Jacob Austin, Michael Isard, Guy {Gur-Ari}, Pengcheng Yin, Toju Duke, Anselm Levskaya, Sanjay Ghemawat, Sunipa Dev, Henryk Michalewski, Xavier Garcia, Vedant Misra, Kevin Robinson, Liam Fedus, Denny Zhou, Daphne Ippolito, David Luan, Hyeontaek Lim, Barret Zoph, Alexander Spiridonov, Ryan Sepassi, David Dohan, Shivani Agrawal, Mark Omernick, Andrew~M. Dai, Thanumalayan~Sankaranarayana Pillai, Marie Pellat, Aitor Lewkowycz, Erica Moreira, Rewon Child, Oleksandr Polozov, Katherine Lee, Zongwei Zhou, Xuezhi Wang, Brennan Saeta, Mark Diaz, Orhan Firat, Michele Catasta, Jason Wei, Kathy {Meier-Hellstern}, Douglas Eck, Jeff Dean, Slav Petrov, and Noah Fiedel.
\newblock {{PaLM}}: {{Scaling Language Modeling}} with {{Pathways}}, October 2022.

\bibitem[Denk and Reisswig(2019)]{Denk2019}
Timo~I. Denk and Christian Reisswig.
\newblock Bertgrid: Contextualized embedding for 2d document representation and understanding.
\newblock \emph{{arXiv} preprint {arXiv}:1909.04948}, 2019.

\bibitem[Devlin et~al.(2019)Devlin, Chang, Lee, and Toutanova]{Devlin2019}
Jacob Devlin, Ming-Wei Chang, Kenton Lee, and Kristina Toutanova.
\newblock {BERT}: Pre-training of deep bidirectional transformers for language understanding.
\newblock In \emph{Proceedings of the 2019 Conference of the North American Chapter of the Association for Computational Linguistics: Human Language Technologies, Volume 1 (Long and Short Papers)}, pages 4171--4186, Minneapolis, Minnesota, June 2019. Association for Computational Linguistics.
\newblock \doi{10.18653/v1/N19-1423}.
\newblock URL \url{https://aclanthology.org/N19-1423}.

\bibitem[Harley et~al.(2015)Harley, Ufkes, and Derpanis]{Harley2015}
Adam~W. Harley, Alex Ufkes, and Konstantinos~G. Derpanis.
\newblock Evaluation of deep convolutional nets for document image classification and retrieval.
\newblock In \emph{International Conference on Document Analysis and Recognition}, number {arXiv}:1502.07058, February 2015.
\newblock URL \url{http://arxiv.org/abs/1502.07058}.
\newblock type: article.

\bibitem[Hochreiter and Schmidhuber(1997)]{Hochreiter1997}
Sepp Hochreiter and Jürgen Schmidhuber.
\newblock Long short-term memory.
\newblock \emph{Neural Computation}, 9\penalty0 (8):\penalty0 1735--1780, November 1997.
\newblock ISSN 0899-7667.
\newblock \doi{10.1162/neco.1997.9.8.1735}.
\newblock URL \url{https://doi.org/10.1162/neco.1997.9.8.1735}.

\bibitem[Huang et~al.(2022)Huang, Lv, Cui, Lu, and Wei]{Huang2022}
Yupan Huang, Tengchao Lv, Lei Cui, Yutong Lu, and Furu Wei.
\newblock {LayoutLMv}3: Pre-training for document {AI} with unified text and image masking, July 2022.
\newblock URL \url{http://arxiv.org/abs/2204.08387}.
\newblock type: article.

\bibitem[Huang et~al.(2019)Huang, Chen, He, Bai, Karatzas, Lu, and Jawahar]{Huang2019}
Zheng Huang, Kai Chen, Jianhua He, Xiang Bai, Dimosthenis Karatzas, Shjian Lu, and C.~V. Jawahar.
\newblock {ICDAR}2019 competition on scanned receipt {OCR} and information extraction.
\newblock In \emph{2019 International Conference on Document Analysis and Recognition ({ICDAR})}, pages 1516--1520, September 2019.
\newblock \doi{10.1109/ICDAR.2019.00244}.
\newblock URL \url{http://arxiv.org/abs/2103.10213}.

\bibitem[Katti et~al.(2018)Katti, Reisswig, Guder, Brarda, Bickel, Höhne, and Faddoul]{Katti2018}
Anoop~R Katti, Christian Reisswig, Cordula Guder, Sebastian Brarda, Steffen Bickel, Johannes Höhne, and Jean~Baptiste Faddoul.
\newblock Chargrid: Towards understanding 2d documents.
\newblock In \emph{Proceedings of the 2018 Conference on Empirical Methods in Natural Language Processing}, pages 4459--4469, Brussels, Belgium, October 2018. Association for Computational Linguistics.
\newblock \doi{10.18653/v1/D18-1476}.
\newblock URL \url{https://aclanthology.org/D18-1476}.

\bibitem[Kim et~al.(2022)Kim, Hong, Yim, Nam, Park, Yim, Hwang, Yun, Han, and Park]{Kim2022}
Geewook Kim, Teakgyu Hong, Moonbin Yim, Jeongyeon Nam, Jinyoung Park, Jinyeong Yim, Wonseok Hwang, Sangdoo Yun, Dongyoon Han, and Seunghyun Park.
\newblock {OCR}-free document understanding transformer, October 2022.
\newblock URL \url{http://arxiv.org/abs/2111.15664}.
\newblock type: article.

\bibitem[Kitaev et~al.(2020)Kitaev, Kaiser, and Levskaya]{Kitaev2020}
Nikita Kitaev, Lukasz Kaiser, and Anselm Levskaya.
\newblock Reformer: The efficient transformer, February 2020.
\newblock URL \url{http://arxiv.org/abs/2001.04451}.
\newblock type: article.

\bibitem[Kolesnikov et~al.(2021)Kolesnikov, Dosovitskiy, Weissenborn, Heigold, Uszkoreit, Beyer, Minderer, Dehghani, Houlsby, Gelly, Unterthiner, and Zhai]{Kolesnikov2021}
Alexander Kolesnikov, Alexey Dosovitskiy, Dirk Weissenborn, Georg Heigold, Jakob Uszkoreit, Lucas Beyer, Matthias Minderer, Mostafa Dehghani, Neil Houlsby, Sylvain Gelly, Thomas Unterthiner, and Xiaohua Zhai.
\newblock An image is worth 16x16 words: Transformers for image recognition at scale.
\newblock In \emph{International Conference on Learning Representations}, 2021.

\bibitem[Lample et~al.(2016)Lample, Ballesteros, Subramanian, Kawakami, and Dyer]{Lample2016}
Guillaume Lample, Miguel Ballesteros, Sandeep Subramanian, Kazuya Kawakami, and Chris Dyer.
\newblock Neural architectures for named entity recognition.
\newblock In \emph{Proceedings of the 2016 Conference of the North American Chapter of the Association for Computational Linguistics: Human Language Technologies}, pages 260--270, San Diego, California, June 2016. Association for Computational Linguistics.
\newblock \doi{10.18653/v1/N16-1030}.
\newblock URL \url{https://aclanthology.org/N16-1030}.

\bibitem[Lewis et~al.(2006)Lewis, Agam, Argamon, Frieder, Grossman, and Heard]{Lewis2006}
D.~Lewis, G.~Agam, S.~Argamon, O.~Frieder, D.~Grossman, and J.~Heard.
\newblock Building a test collection for complex document information processing.
\newblock In \emph{Proceedings of the 29th annual international {ACM} {SIGIR} conference on Research and development in information retrieval}, {SIGIR} '06, pages 665--666, New York, {NY}, {USA}, August 2006. Association for Computing Machinery.
\newblock ISBN 9781595933690.
\newblock \doi{10.1145/1148170.1148307}.
\newblock URL \url{https://doi.org/10.1145/1148170.1148307}.

\bibitem[Li et~al.(2020)Li, Xu, Cui, Huang, Wei, Li, and Zhou]{Li2020}
Minghao Li, Yiheng Xu, Lei Cui, Shaohan Huang, Furu Wei, Zhoujun Li, and Ming Zhou.
\newblock {DocBank}: A benchmark dataset for document layout analysis.
\newblock In \emph{Proceedings of the 28th International Conference on Computational Linguistics}, pages 949--960, Barcelona, Spain (Online), December 2020. International Committee on Computational Linguistics.
\newblock \doi{10.18653/v1/2020.coling-main.82}.
\newblock URL \url{https://aclanthology.org/2020.coling-main.82}.

\bibitem[Liu et~al.(2019)Liu, Ott, Goyal, Du, Joshi, Chen, Levy, Lewis, Zettlemoyer, and Stoyanov]{Liu2019}
Yinhan Liu, Myle Ott, Naman Goyal, Jingfei Du, Mandar Joshi, Danqi Chen, Omer Levy, Mike Lewis, Luke Zettlemoyer, and Veselin Stoyanov.
\newblock {RoBERTa}: A robustly optimized {BERT} pretraining approach, July 2019.
\newblock URL \url{http://arxiv.org/abs/1907.11692}.
\newblock type: article.

\bibitem[Mathew et~al.(2021)Mathew, Karatzas, and Jawahar]{Mathew2021}
Minesh Mathew, Dimosthenis Karatzas, and C.~V. Jawahar.
\newblock {DocVQA}: A dataset for {VQA} on document images.
\newblock In \emph{Winter Conference on Applications of Computer Vision}, pages 2200--2209, 2021.
\newblock URL \url{https://openaccess.thecvf.com/content/WACV2021/html/Mathew_DocVQA_A_Dataset_for_VQA_on_Document_Images_WACV_2021_paper.html}.

\bibitem[Mikolov et~al.(2013)Mikolov, Sutskever, Chen, Corrado, and Dean]{Mikolov2013}
Tomas Mikolov, Ilya Sutskever, Kai Chen, Greg~S Corrado, and Jeff Dean.
\newblock Distributed representations of words and phrases and their compositionality.
\newblock In \emph{Advances in Neural Information Processing Systems}, volume~26. Curran Associates, Inc., 2013.
\newblock URL \url{https://papers.nips.cc/paper/2013/hash/9aa42b31882ec039965f3c4923ce901b-Abstract.html}.

\bibitem[Pennington et~al.(2014)Pennington, Socher, and Manning]{Pennington2014}
Jeffrey Pennington, Richard Socher, and Christopher Manning.
\newblock {GloVe}: Global vectors for word representation.
\newblock In \emph{Proceedings of the 2014 Conference on Empirical Methods in Natural Language Processing ({EMNLP})}, pages 1532--1543, Doha, Qatar, October 2014. Association for Computational Linguistics.
\newblock \doi{10.3115/v1/D14-1162}.
\newblock URL \url{https://aclanthology.org/D14-1162}.

\bibitem[Peters et~al.(2018)Peters, Neumann, Iyyer, Gardner, Clark, Lee, and Zettlemoyer]{Peters2018}
Matthew~E. Peters, Mark Neumann, Mohit Iyyer, Matt Gardner, Christopher Clark, Kenton Lee, and Luke Zettlemoyer.
\newblock Deep contextualized word representations.
\newblock In \emph{Proceedings of the 2018 Conference of the North American Chapter of the Association for Computational Linguistics: Human Language Technologies, Volume 1 (Long Papers)}, pages 2227--2237, New Orleans, Louisiana, June 2018. Association for Computational Linguistics.
\newblock \doi{10.18653/v1/N18-1202}.
\newblock URL \url{https://aclanthology.org/N18-1202}.

\bibitem[Powalski et~al.(2021)Powalski, Borchmann, Jurkiewicz, Dwojak, Pietruszka, and Pałka]{Powalski2021}
Rafał Powalski, Lukasz Borchmann, Dawid Jurkiewicz, Tomasz Dwojak, Michał Pietruszka, and Gabriela Pałka.
\newblock Going full-{TILT} boogie on document understanding with text-image-layout transformer, July 2021.
\newblock URL \url{http://arxiv.org/abs/2102.09550}.
\newblock type: article.

\bibitem[Qin et~al.(2022)Qin, Sun, Deng, Li, Wei, Lv, Yan, Kong, and Zhong]{Qin2022}
Zhen Qin, Weixuan Sun, Hui Deng, Dongxu Li, Yunshen Wei, Baohong Lv, Junjie Yan, Lingpeng Kong, and Yiran Zhong.
\newblock {cosFormer}: Rethinking softmax in attention.
\newblock In \emph{International Conference on Learning Representations}, number {arXiv}:2202.08791, February 2022.
\newblock \doi{10.48550/arXiv.2202.08791}.
\newblock URL \url{http://arxiv.org/abs/2202.08791}.
\newblock type: article.

\bibitem[Radford et~al.(2019)Radford, Wu, Child, Luan, Amodei, and Sutskever]{Radford2019}
Alec Radford, Jeff Wu, Rewon Child, David Luan, Dario Amodei, and Ilya Sutskever.
\newblock Language models are unsupervised multitask learners.
\newblock 2019.

\bibitem[Shaw et~al.(2018)Shaw, Uszkoreit, and Vaswani]{Shaw2018}
Peter Shaw, Jakob Uszkoreit, and Ashish Vaswani.
\newblock Self-attention with relative position representations.
\newblock In \emph{Proceedings of the 2018 Conference of the North American Chapter of the Association for Computational Linguistics: Human Language Technologies, Volume 2 (Short Papers)}, pages 464--468, New Orleans, Louisiana, June 2018. Association for Computational Linguistics.
\newblock \doi{10.18653/v1/N18-2074}.
\newblock URL \url{https://aclanthology.org/N18-2074}.

\bibitem[Vaswani et~al.(2017)Vaswani, Shazeer, Parmar, Uszkoreit, Jones, Gomez, Kaiser, and Polosukhin]{Vaswani2017}
Ashish Vaswani, Noam Shazeer, Niki Parmar, Jakob Uszkoreit, Llion Jones, Aidan~N Gomez, Lukasz Kaiser, and Illia Polosukhin.
\newblock Attention is all you need.
\newblock In \emph{Advances in Neural Information Processing Systems}, volume~30. Curran Associates, Inc., 2017.
\newblock URL \url{https://papers.nips.cc/paper/2017/hash/3f5ee243547dee91fbd053c1c4a845aa-Abstract.html}.

\bibitem[Wang et~al.(2018)Wang, Singh, Michael, Hill, Levy, and Bowman]{Wang2018}
Alex Wang, Amanpreet Singh, Julian Michael, Felix Hill, Omer Levy, and Samuel Bowman.
\newblock {GLUE}: A multi-task benchmark and analysis platform for natural language understanding.
\newblock In \emph{Proceedings of the 2018 {EMNLP} Workshop {BlackboxNLP}: Analyzing and Interpreting Neural Networks for {NLP}}, pages 353--355, Brussels, Belgium, November 2018. Association for Computational Linguistics.
\newblock \doi{10.18653/v1/W18-5446}.
\newblock URL \url{https://aclanthology.org/W18-5446}.

\bibitem[Wang et~al.(2020)Wang, Li, Khabsa, Fang, and Ma]{Wang2020}
Sinong Wang, Belinda~Z. Li, Madian Khabsa, Han Fang, and Hao Ma.
\newblock Linformer: Self-attention with linear complexity, June 2020.
\newblock URL \url{http://arxiv.org/abs/2006.04768}.
\newblock type: article.

\bibitem[Xu et~al.(2021)Xu, Xu, Lv, Cui, Wei, Wang, Lu, Florencio, Zhang, Che, Zhang, and Zhou]{Xu2021}
Yang Xu, Yiheng Xu, Tengchao Lv, Lei Cui, Furu Wei, Guoxin Wang, Yijuan Lu, Dinei Florencio, Cha Zhang, Wanxiang Che, Min Zhang, and Lidong Zhou.
\newblock {LayoutLMv}2: Multi-modal pre-training for visually-rich document understanding.
\newblock In \emph{Proceedings of the 59th Annual Meeting of the Association for Computational Linguistics and the 11th International Joint Conference on Natural Language Processing (Volume 1: Long Papers)}, pages 2579--2591, Online, August 2021. Association for Computational Linguistics.
\newblock \doi{10.18653/v1/2021.acl-long.201}.
\newblock URL \url{https://aclanthology.org/2021.acl-long.201}.

\bibitem[Xu et~al.(2020)Xu, Li, Cui, Huang, Wei, and Zhou]{Xu2020}
Yiheng Xu, Minghao Li, Lei Cui, Shaohan Huang, Furu Wei, and Ming Zhou.
\newblock {LayoutLM}: Pre-training of text and layout for document image understanding.
\newblock In \emph{Proceedings of the 26th {ACM} {SIGKDD} International Conference on Knowledge Discovery \& Data Mining}, {KDD} '20, pages 1192--1200, New York, {NY}, {USA}, August 2020. Association for Computing Machinery.
\newblock ISBN 9781450379984.
\newblock \doi{10.1145/3394486.3403172}.
\newblock URL \url{https://doi.org/10.1145/3394486.3403172}.

\bibitem[Yu et~al.(2020)Yu, Lu, Qi, Gong, and Xiao]{Yu2020}
Wenwen Yu, Ning Lu, Xianbiao Qi, Ping Gong, and Rong Xiao.
\newblock {PICK}: Processing key information extraction from documents using improved graph learning-convolutional networks, July 2020.
\newblock URL \url{http://arxiv.org/abs/2004.07464}.
\newblock type: article.

\bibitem[Zaheer et~al.(2020)Zaheer, Guruganesh, Dubey, Ainslie, Alberti, Ontanon, Pham, Ravula, Wang, Yang, and Ahmed]{Zaheer2020}
Manzil Zaheer, Guru Guruganesh, Kumar~Avinava Dubey, Joshua Ainslie, Chris Alberti, Santiago Ontanon, Philip Pham, Anirudh Ravula, Qifan Wang, Li~Yang, and Amr Ahmed.
\newblock Big bird: Transformers for longer sequences.
\newblock In \emph{Advances in Neural Information Processing Systems}, volume~33, pages 17283--17297. Curran Associates, Inc., 2020.
\newblock URL \url{https://proceedings.neurips.cc/paper/2020/hash/c8512d142a2d849725f31a9a7a361ab9-Abstract.html}.

\end{thebibliography}
\end{document}